\documentclass{article}
\usepackage{spconf,amsmath,graphicx}
\usepackage{hyperref}
\hypersetup{
    colorlinks = true,
    urlcolor = blue,
}

\title{Weakly supervised segmentation of cracks on solar cells using normalized $\mathbf{L_p}$ norm}
\name{Martin Mayr, Mathis Hoffmann, Andreas Maier, Vincent Christlein}
\address{Pattern Recognition Lab, Friedrich-Alexander University Erlangen-Nuremberg, Germany\\
\{martin.mayr, mathis.hoffmann, andreas.maier, vincent.christlein\}@fau.de}
\begin{document}
\maketitle
\begin{abstract}
\vspace{0.5cm}
Photovoltaic is one of the most important renewable energy sources for dealing with world-wide steadily increasing energy consumption. This raises the demand for fast and scalable automatic quality management during production and operation. However, the detection and segmentation of cracks on electroluminescence (EL) images of mono- or polycrystalline solar modules is a challenging task. In this work, we propose a weakly supervised learning strategy that only uses image-level annotations to obtain a method that is capable of segmenting cracks on EL images of solar cells. We use a modified ResNet-50 to derive a segmentation from network activation maps. We use defect classification as a surrogate task to train the network. To this end, we apply normalized $L_p$ normalization to aggregate the activation maps into single scores for classification. In addition, we provide a study how different parameterizations of the normalized $L_p$ layer affect the segmentation performance. This approach shows promising results for the given task. However, we think that the method has the potential to solve other weakly supervised segmentation problems as well.
\end{abstract}
\begin{keywords}
crack detection, weakly supervised semantic segmentation, EL imaging, solar cell, normalized $L_p$ norm
\end{keywords}

\section{Introduction}
The need of renewable energy sources is rapidly growing in times of climate change and world wide increasing energy consumption. Already today, a major part of renewable energy is obtained by photovoltaics~(PV)~\cite{nations2018}.
The important role of solar power in the transition from conventional to renewable power sources enforces a thorough assessment of the quality of PV modules from production to operation. Therefore, an automatic detection and classification of certain defects is essential.

\begin{figure}[t]
    \centering
    \includegraphics[width=0.35\textwidth]{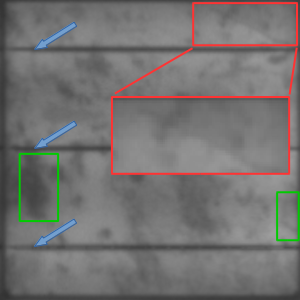}
    \caption{EL image of a polycrystalline solar cell with a crack in the right upper corner. Red box shows the area where the crack is located. Blue arrows point to the busbars. Green boxes show noisy areas, which are no crack regions.}
    \label{fig:example}
\end{figure}

In this work, we propose a crack segmentation pipeline that is applied on different types of solar cells. Especially, the grainy and noisy structure of polycrystalline solar cells renders an automatic segmentation, classification or even localization of defects challenging (cf. Fig.~\ref{fig:example}). We use a publicly available dataset of electroluminescence (EL) images of solar cells. Since segmentation masks are unavailable and cost intensive to obtain, we resort to a weakly supervised segmentation approach using deep learning. This way, we learn a rough segmentation of the cell from image-level labels only. In addition, we systematically analyze the use of the normalized $L_p$ norm with different values of $p$ for the task of weakly supervised segmentation.

The main contributions of this work are: 
\begin{itemize}
    \item We developed a model that is capable of segmenting cracks on EL images of mono- or polycrystalline solar modules with one forward path.
    \item We analyzed the effect of different normalized ${L_p}$ norms in case of weakly supervised semantic segmentation.  
\end{itemize}


\begin{figure*}[t]
    \centering
    \includegraphics[width=1\textwidth]{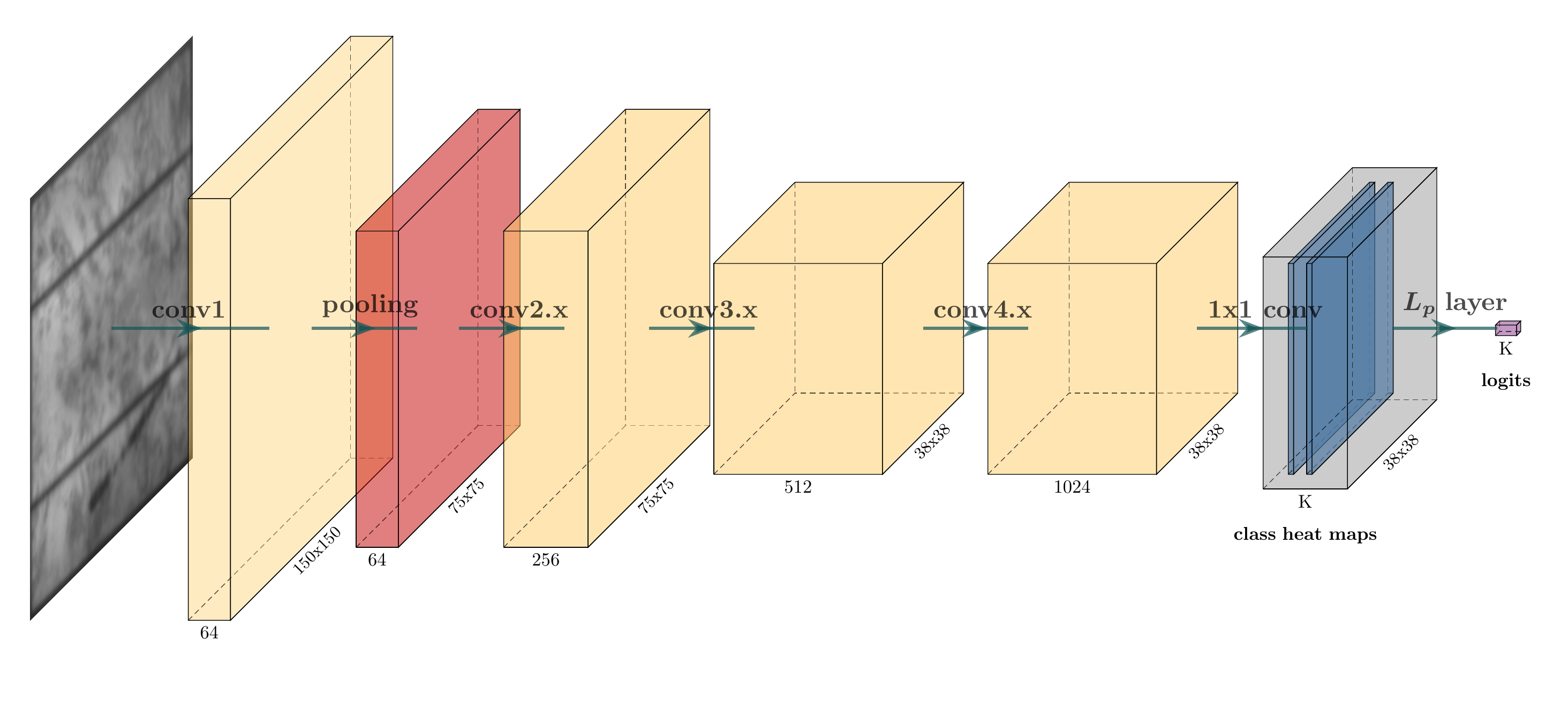}
    \caption{Architecture of modified ResNet50. In contrast to the original ResNet50, this version is just downsampling in conv1 and conv3.x to retain more precise spatial information. Downsampling is performed via stride equal to 2 for the first convolutional layer of the scaled residual block.}
    \label{fig:resnet}
\end{figure*}

\section{Related Work}
There are several algorithms which have been proposed to solve the problem of segmenting EL images of solar cells. In many cases, over-rejection of small cracks is a problem and, on the other hand, defects like dislocations are easily interpreted as cracks leading to a large number of false positives
~\cite{anwar2014micro, lin2011automatic, spataru2016automatic, stromer2019eCS, tsai2012defect}. 
Since these works have shown that this problem is hard to solve with classical image processing methods, we apply a deep learning approach to solve it in a data driven way.

Weakly supervised segmentation with image-level labels has only rarely been addressed in research. As one of the pioneers, Oquab~et.~al.~\cite{Oquab_2015_CVPR} recognized the potential of this learning strategy. They used a fully convolutional architecture with five convolutional blocks and added a global max-pooling layer at the end of the network. This approach results in a good localization of objects on the Pascal VOC~12 and MS~COCO dataset. Later, Wei~et.~al.~\cite{wei2018revisiting} extended the convolutional layers with different dilation rates to connect discriminative object regions with non-discriminative object regions. Huang~et.~al.~\cite{huang2018weakly} localize the seeds with a classification network and  use these information cues for seeded region growing (SRG). In addition this SRG approach is extended with a segmentation network and a special seeding loss function. The results are promising. On the other hand, approaches based on Oquab~et.~al.~\cite{Oquab_2015_CVPR} are attractive due to their simplicity and computational efficiency.
In our architecture we use the normalized $L_p$ norm described by Gulcehre~et.~al.~\cite{gulcehre2014} as a global pooling layer separatly for each feature map. A similar pooling strategy has been used in the context of image retrieval by Radenović ~et.~al.~\cite{radenovic2016cnn}.
\begin{figure*}[htb]
    \centering
    \begin{minipage}[b]{0.16\linewidth}
      \centering
      \centerline{\includegraphics[width=1\textwidth]{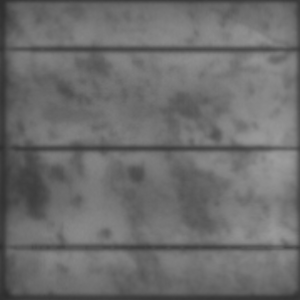}}
      \centerline{(a) Original EL image}\medskip
    \end{minipage}
    \begin{minipage}[b]{0.16\linewidth}
      \centering
      \centerline{\includegraphics[width=1\textwidth]{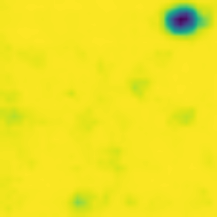}}
      \centerline{(b) $L_1$}\medskip
    \end{minipage}
    \begin{minipage}[b]{0.16\linewidth}
      \centering
      \centerline{\includegraphics[width=1\textwidth]{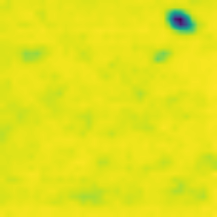}}
      \centerline{(c) $L_2$}\medskip
    \end{minipage}
    \begin{minipage}[b]{0.16\linewidth}
      \centering
      \centerline{\includegraphics[width=1\textwidth]{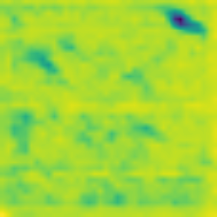}}
      \centerline{(d) $L_3$}\medskip
    \end{minipage}
    \begin{minipage}[b]{0.16\linewidth}
      \centering
      \centerline{\includegraphics[width=1\textwidth]{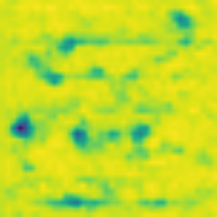}}
      \centerline{(e) $L_4$}\medskip
    \end{minipage}
    \begin{minipage}[b]{0.16\linewidth}
      \centering
      \centerline{\includegraphics[width=1\textwidth]{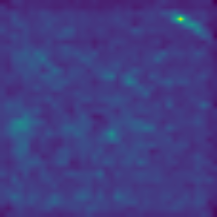}}
      \centerline{(f) $L_{\infty}$}\medskip
    \end{minipage}
    
    \caption{Original EL image and heat maps of crack layer for different normalized $L_p$ norms. In the heat maps the dark areas are low values and the bright values are high values.}
    \label{fig:lp_results}
\end{figure*}
\section{Materials \& Methods}
\subsection{Training data and Preprocessing}
The dataset used for this work\footnote{\url{https://github.com/zae-bayern/elpv-dataset}} consists of 2426 electroluminescence 8-bit grayscale images of solar cells with a resolution of $300 \times 300$ pixels per image~\cite{Buerhop2018,Deitsch2018a,Deitsch2018b}. These images were extracted from 44 different solar modules. It includes expert annotations of the probability that a cell contains any kind of defect. We manually added image-level annotations of the defect type, since we are interested in cracks only. The dataset is class-imbalanced with much more non-crack samples than crack samples. It contains monocrystalline as well as polycristalline cells.

In Fig.~\ref{fig:example}, an EL image of a polycrystalline solar cell is shown. The blue arrows point to the three vertical busbars which are visible on most of the samples and should not be segmented as crack regions. Since the configuration of busbars varies from cell type to cell type, the method needs to ignore the busbars reliably. Especially for polychristalline cells, there are instances, where the structure of the silicon wafers is similar to the apparence of a crack (green boxes). Of course, neither of them should be segmented as cracks.

\subsection{Architecture}

We start with a general classification architecture and modify it to obtain the segmentation result. The basis for this approach is a residual network with 50 layers~\cite{He_2016_CVPR}. Due to the limited amount of data, best results are achieved with transfer learning. The weights are initialized with pretrained weights trained on ImageNet~\cite{imagenet_cvpr09}. We use categorical cross entropy as loss function. 

We apply four modifications to adapt the network to the given task:
\begin{enumerate}
    \item To preserve spatial information throughout the forward path and increase the resolution of the segmentation masks, we remove the fully connected layer, the average pooling layer and one of the bottleneck blocks.
    \item The last bottleneck block is changed to have a stride of $1$, such that the resolution is not affected. To this end, the stride in the first convolutional layer is set to $1$. To allow using the pretrained weights, we set the dilation rate of the following convolution to $2$.
    \item A $1 \times 1$ convolutional layer with two feature maps is added to adapt the number of activation maps such that it matches the number of classes. In our case, the first feature map shows the activations for 'crack' and the other one shows the activations for 'non-crack'. 
    \item A normalized $L_p$ norm layer is added to aggregate the activation maps into a single value for classification.
\end{enumerate}

The input data from the feature maps is flattened to a finite set of input signals represented as vector $\vec{x}$ with $N$ elements. This vector is fed into the normalized $L_p$ norm, which is defined as
\begin{equation}\label{eq:lp_norm}
    L_p(\vec{x}) = (\frac{1}{N}\Sigma_{i=1}^N |y_i| ^{p})^\frac{1}{p} \quad ,
\end{equation}
where $x_i$ denote the $N$ pixels of the corresponding activation map $\mathbf{Y}$.

Certain normalized $L_p$ norms are equal to some types of pooling: For example, $L_1$ is identical to average pooling
and $L_{\infty}$ is equal to max pooling
\cite{Boureau2010, gulcehre2014}. 

\subsection{Segmentation task}

The classification problem is used as a surrogate task to train the $1 \times 1$ convolutional layer. At test time, this is a segmentation problem. Hence, we are only interested in the activations of conv5. These activations highlight the important regions during inference. The resolution of both heat maps is 38 $\times$ 38 due to the downsampling of the ResNet-50. For the segmentation task, we only use the map that corresponds to the class 'crack'. In addition, we only apply the segmentation, when the classifications states that there is a crack on the input image. Then, the segmentation is simply done by pixel-wise comparing the heat map value with the half of the maximum value of the heat map
\begin{equation}
    segment(y_i) =
  \begin{cases}
    1, & \text{for } y_i >  \frac{\max(\mathbf{Y})}{2} \\
    0, & \text{else} 
  \end{cases}
  \label{eq:seg}
\end{equation}
If the compared value is above it is a crack region, otherwise it's background, shown in equation \ref{eq:seg}. The threshold is computed dynamically, since the activations are scaled arbitrarily.

Note that the actual segmentation is not the main purpose of this paper and a better segmentation can certainly be achieved by an advanced method.

\section{Results}

First, we present classification results for different normalized $L_p$ norms to show how the surrogate task performs using different values of $p$. Then, the heat maps and segmentation results are qualitatively evaluated to check whether good classification performances result in good segmentation results.

\subsection{Classification}

\begin{table}[t!]
\begin{tabular}{|c|c|c|c|c|c|c|c|}
\hline
\textbf{Norm} & $\mathbf{L_1}$ & $\mathbf{L_2}$ & $\mathbf{L_3}$ & $\mathbf{L_4}$ & $\mathbf{L_5}$ & $\mathbf{L_9}$ & $\mathbf{L_{\infty}}$ \\
\hline
\textbf{Precision} & 0.90 & 0.82 & 0.87 & 0.83 & 0.54 & 0.58 & 0.77 \\
\textbf{Recall} & 0.77 & 0.78 & 0.80 & 0.67 & 0.70 & 0.63 & 0.77 \\
\textbf{f1 score} & 0.83 & 0.80 & 0.83 & 0.74 & 0.61 & 0.61 & 0.77 \\
\hline
\textbf{Accuracy} & 0.94 & 0.92 & 0.94 & 0.91 & 0.82 & 0.84 & 0.91\\  
\hline
\end{tabular}
\caption{Classification results for different $L_p$ norms over the elpv-dataset. Precision, recall and f1 score measured on class 'crack'. Accuracy measured on all test data.}
\label{t:classification}
\end{table}

In Tab.~\ref{t:classification}, classification results for the class 'crack' are shown for different values of $p$. It turns out that
$L_1$ and $L_3$ have the best accuracy and f1 score during classification of the test samples. $L_3$ has a slightly lower precision but higher recall than $L_1$. The norms $L_2$, $L_4$ and $L_{\infty}$ have the same accuracy, but the f1 score differs strongly between them. 
The low f1 score of class 'crack' of $L_4$ shows that the model favors the 'non-crack' class. The norms between $L_5$ and $L_9$ have low f1 scores and are therefore not taken into further consideration.
Based on these classification results, $L_1$ and $L_3$ perform best.

\subsection{Segmentation}

Fig.~\ref{fig:lp_results} shows the results of one sample for different normalized $L_p$ norms. These results differ from the classification results. $L_1$ localizes the cracks very well, but it can't preserve the structure and leads to a blurry result. $L_2$ has a similar behavior but with a finer clue for the location. The activation map of $L_3$ gets the shape quite well. From $L_4$ to $L_9$ there is no region shown which can be interpreted as a crack.
By contrast the activation map of $L_{\infty}$ describes the crack very sharply. Hence it serves as a basis for the segmentation.

Interestingly, for $L_1$ to $L_4$, the important regions contain low values in the activation maps and the mean of the activation map of class 'crack' is very high. 
For larger values of $p$ this behavior is inverted, so that the important regions are represented as high values in the activation map.
This is certainly explained by the fact that cracks cover only a small region of the entire image. To obtain a good classification score, a large number of pixels needs to be high in case of small $p$, which is achieved by the inversion of intensities.

\begin{figure}[t!]
    \centering
    \begin{minipage}[b]{0.32\linewidth}
      \centering
      \centerline{\includegraphics[width=1\textwidth]{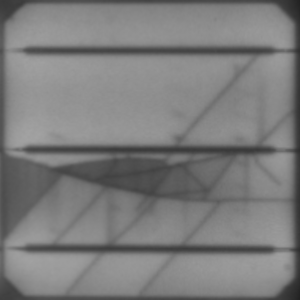}}
      \centerline{(a) Input image}\medskip
    \end{minipage}
    \begin{minipage}[b]{0.32\linewidth}
      \centering
      \centerline{\includegraphics[width=1\textwidth]{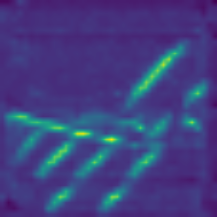}}
      \centerline{(b) Heat map}\medskip
    \end{minipage}
    \begin{minipage}[b]{0.32\linewidth}
      \centering
      \centerline{\includegraphics[width=1\textwidth]{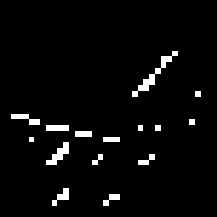}}
      \centerline{(c) Segmentation}\medskip
    \end{minipage}
    \begin{minipage}[b]{0.32\linewidth}
      \centering
      \centerline{\includegraphics[width=1\textwidth]{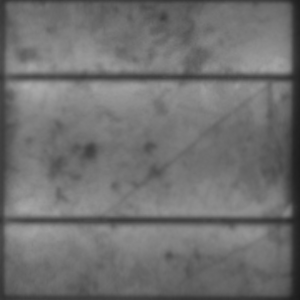}}
      \centerline{(d) Input image}\medskip
    \end{minipage}
    \begin{minipage}[b]{0.32\linewidth}
      \centering
      \centerline{\includegraphics[width=1\textwidth]{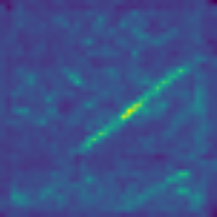}}
      \centerline{(e) Heat map}\medskip
    \end{minipage}
    \begin{minipage}[b]{0.32\linewidth}
      \centering
      \centerline{\includegraphics[width=1\textwidth]{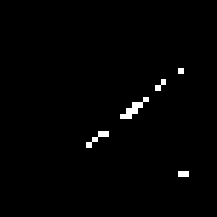}}
      \centerline{(f) Segmentation}\medskip
    \end{minipage}
    \caption{Results for input images of mono- and polycrystalline solar cells with normalized $L_{\infty}$ norm}
    \label{fig:mono_results}
\end{figure}

In Fig.~\ref{fig:mono_results} the difference between heat maps for mono- and polycrystalline cells is highlighted. For EL images of monocrystalline cells, there is no noisy background and the structure of the crack is preserved nicely. In contrast, the heat maps for polycrystalline cells have a noisy background. Nevertheless, our proposed segmentation method gives good results in any case.

\section{Discussion}

The results show that the choice of the best norm for a given problem is crucial for the quality of the weakly supervised segmentation results. While lower norms lead to blurred results, higher norms give very sharp results. From this behavior, we conclude that lower norms might be better suited for segmenting large, blob-like objects, whereas higher norms may be preferred for tiny objects with sharp edges.

For the given problem of segmenting cracks on solar cells, we see that $L_{\infty}$ performs best, but $L_{3}$ could be a good choice as well, if the foreground-background inversion is taken into account. In addition, we show that segmentation performance is not strictly related to classification performance.


\section{Conclusion}
In this paper, we propose an effective technique to get a coarse segmentation of cracks from EL images of mono- and polycrystalline solar cells. We show that given only a small number of training samples and only image-level annotations, the method is able to reliably segment cracks on solar cells with just one forward pass through the network. We are confident, that this is an important contribution to an effective production and maintanance of solar power plants.

This application-oriented result is supplemented by a comprehensive study of the effect of normalized $L_p$ normalization with different values of $p$ on the result of weakly supervised segmentation methods. Thus, this method can serve as a basis for the design of new weakly supervised segmentation procedures.

\vspace{.5cm}

\textbf{Acknowledgement} \vspace{.1cm} \\
We gratefully acknowledge the Federal Ministry for Economic Affairs and Energy (BMWi: Grant No. 0324286D, iPV4.0) for funding.

\bibliographystyle{abbrv}
\bibliography{references}

\end{document}